\documentclass[runningheads]{llncs}

\usepackage{eccv}

\usepackage{eccvabbrv}

\usepackage{graphicx}
\usepackage{booktabs}

\usepackage[accsupp]{axessibility}  

\usepackage{hyperref}

\usepackage{orcidlink}

\usepackage{latexsym}
\usepackage{caption}
\usepackage{amsmath}
\usepackage{bm}
\usepackage{booktabs} 
\usepackage{multirow} 
\usepackage{graphicx} 
\usepackage[table,xcdraw]{xcolor} 
\usepackage{soul}
\usepackage{tabularx}
\usepackage{xcolor}
\usepackage{listings}
\usepackage{float}

\newfloat{algorithm}{tbp}{loa}
\floatname{algorithm}{Algorithm}
\newcounter{algorithmicline}
\newenvironment{algorithmic}[1][0]{%
  \begin{list}{}{\setlength{\leftmargin}{2.5em}\setlength{\itemsep}{0pt}}%
  \setcounter{algorithmicline}{0}%
}{\end{list}}
\newcommand{\STATE}{\refstepcounter{algorithmicline}\item[\scriptsize\arabic{algorithmicline}:]}
\newcommand{\REQUIRE}{\item[\textbf{Require:}]}
\newcommand{\ENSURE}{\item[\textbf{Ensure:}]}
\newcommand{\FOR}[1]{\STATE \textbf{for} #1 \textbf{do}}
\newcommand{\ENDFOR}{\STATE \textbf{end for}}
\newcommand{\IF}[1]{\STATE \textbf{if} #1 \textbf{then}}
\newcommand{\ELSE}{\STATE \textbf{else}}
\newcommand{\ENDIF}{\STATE \textbf{end if}}
\newcommand{\RETURN}{\STATE \textbf{return} }
\newcommand{\TO}{\textbf{to}}
\newcommand{\pad}{\hphantom{\scriptsize~$\uparrow$0.00}}
\newcommand{\res}[2]{%
    \makebox[0pt][r]{#1\hspace{0.4em}}
    \rlap{\footnotesize\color{GainGreen}$\uparrow$#2}
}

\usepackage{tcolorbox}
\definecolor{BgPerception}{rgb}{0.92, 0.96, 1.0}  
\definecolor{BgReasoning}{rgb}{0.94, 1.0, 0.94}   
\definecolor{BgRobustness}{rgb}{1.0, 0.96, 0.92}  
\definecolor{HeaderGray}{gray}{0.92}
\definecolor{OursHighlight}{rgb}{0.96, 0.98, 1.0}
\definecolor{GainGreen}{rgb}{0.0, 0.6, 0.0}
\definecolor{eclipseBlue}{RGB}{42,0.0,255}
\definecolor{eclipseGreen}{RGB}{63,127,95}
\definecolor{eclipsePurple}{RGB}{127,0,85}
\definecolor{backcolour}{RGB}{245,245,245} 

\lstdefinelanguage{json}{
    basicstyle=\ttfamily\small,
    commentstyle=\color{eclipseGreen}\ttfamily,
    stringstyle=\color{eclipseBlue},
    numbers=left,
    numberstyle=\scriptsize\color{gray},
    stepnumber=1,
    numbersep=8pt,
    showstringspaces=false,
    breaklines=true,
    frame=lines,
    backgroundcolor=\color{backcolour},
    literate=
     *{0}{{{\color{eclipsePurple}0}}}{1}
      {1}{{{\color{eclipsePurple}1}}}{1}
      {2}{{{\color{eclipsePurple}2}}}{1}
      {3}{{{\color{eclipsePurple}3}}}{1}
      {4}{{{\color{eclipsePurple}4}}}{1}
      {5}{{{\color{eclipsePurple}5}}}{1}
      {6}{{{\color{eclipsePurple}6}}}{1}
      {7}{{{\color{eclipsePurple}7}}}{1}
      {8}{{{\color{eclipsePurple}8}}}{1}
      {9}{{{\color{eclipsePurple}9}}}{1}
      {:}{{{\color{black}{:}}}}{1}
      {,}{{{\color{black}{,}}}}{1}
      {\{}{{{\color{black}{\{}}}}{1}
      {\}}{{{\color{black}{\}}}}}{1}
      {[}{{{\color{black}{[}}}}{1}
      {]}{{{\color{black}{]}}}}{1},
      morecomment=[l]{//}, 
}
\begin{document}

\title{LaViT: Aligning Latent Visual Thoughts for Multi-modal Reasoning} 


\author{Linquan Wu\textsuperscript{*}\inst{1}\orcidlink{0009-0000-6594-625X} \and
Tianxiang Jiang\textsuperscript{*}\inst{2}\orcidlink{0009-0001-3957-9748} \and
Yifei Dong\inst{3}\orcidlink{0009-0004-9276-1399} \and
Haoyu Yang\inst{4} \and
Fengji Zhang\inst{1}\orcidlink{0000-0002-0965-417X} \and
Shichang Meng\inst{1}\orcidlink{0009-0008-6971-2707} \and
Ai Xuan\inst{1}\orcidlink{0000-0002-0801-3549} \and
Linqi Song\inst{1}\orcidlink{0000-0003-2756-4984} \and
Jacky Keung\inst{1}\orcidlink{0000-0002-3803-9600}}

\authorrunning{L.~Wu, T.~Jiang et al.}

\institute{City University of Hong Kong \and
University of Science and Technology of China \and
Utrecht University \and
University of Electronic Science and Technology of China}

\maketitle
\begingroup
\renewcommand{\thefootnote}{*}\footnotetext{Equal contribution.}
\endgroup

\begin{abstract}
Current multimodal latent reasoning often relies on external supervision (e.g., auxiliary images), ignoring intrinsic visual attention dynamics. In this work, we identify a critical \textbf{Perception Gap} in distillation: student models frequently mimic a teacher's textual output while attending to fundamentally divergent visual regions, effectively relying on language priors rather than grounded perception. To bridge this, we propose \textbf{LaViT}, a framework that aligns \textit{latent visual thoughts} rather than static embeddings. LaViT compels the student to autoregressively reconstruct the teacher's visual semantics and attention trajectories prior to text generation, employing a \textit{curriculum sensory gating} mechanism to prevent shortcut learning. Extensive experiments show that LaViT significantly enhances visual grounding, achieving up to \textbf{+16.9\%} gains on complex reasoning tasks and enabling a compact 3B model to outperform larger open-source variants and proprietary models like GPT-4o.   
\end{abstract}

\section{Introduction}

Multimodal Large Language Models (MLLMs) have advanced rapidly in recent years~\cite{bai2025qwen3vltechnicalreport,comanici2025gemini,seed2025seed1_5vl}, 
early multimodal reasoning models primarily \emph{thinking about images}, captioning visual inputs into text and reasoning mainly in the language space via explicit chains of thought~\cite{huang2025vision,yang2025r1,shen2025vlm}. Recent work instead emphasizes \emph{thinking with images}, more tightly integrating visual evidence into the reasoning process~\cite{hu2024visual,zheng2025deepeyes,su2025thinking}, leading to improved performance on complex visual reasoning tasks~\cite{fu2024blink,wu2024v,zhang2024humaneval,wang2025monosropenvocabularyspatialreasoning}

\begin{figure*}[t]  
    \centering
    \vspace{-3mm}
    \includegraphics[width=\textwidth]{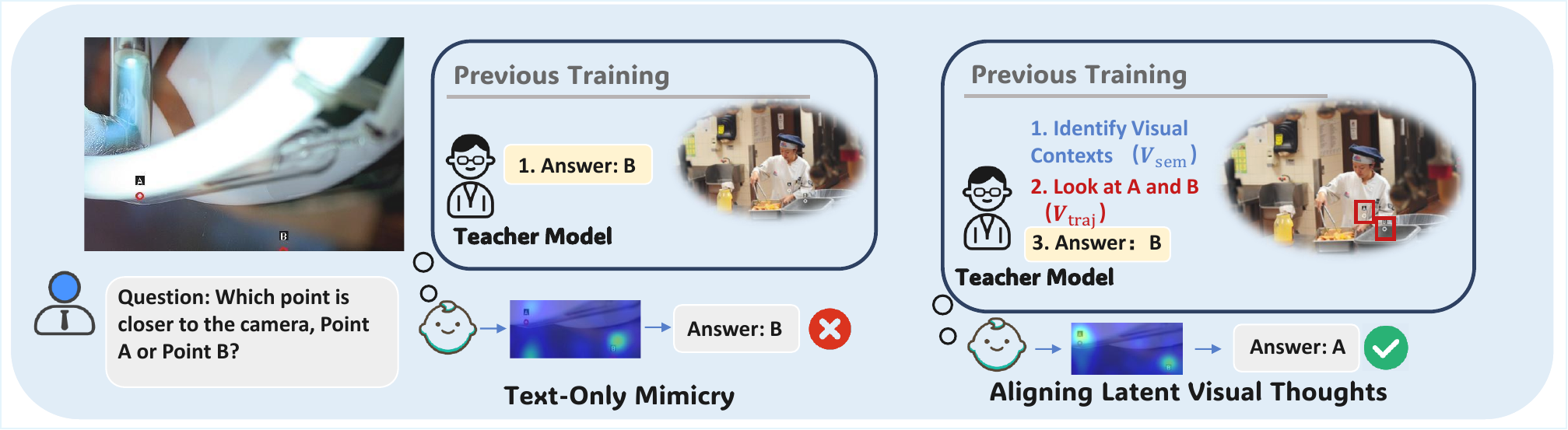}
    \vspace{-7mm}
    \caption{Conceptual Illustration of Our Proposed Method LaViT.}
    \vspace{-3mm}
\label{fig:illustartion}
\end{figure*}

Subsequently, latent reasoning~\cite{hao2024training} has emerged as a complementary direction, compressing intermediate reasoning into continuous hidden states rather than explicit CoT. This paradigm has been extended to MLLMs to model abstract visual thoughts within latent tokens~\cite{yang2025machine, li2025latent}. While effective, existing methods largely rely on \emph{manually designed} visual supervision, such as auxiliary images or annotated regions, leaving intrinsic visual attention dynamics during reasoning unexplored.

These limitations motivate the use of knowledge distillation~\cite{hinton2015distilling} as a lens to analyze and transfer visual reasoning behaviors in MLLMs. Distilling a high-capacity teacher into a compact student enables us to probe not only \emph{what} knowledge is transferred, but also \emph{how} visual reasoning is internally realized. However, existing multimodal distillation methods mainly align final textual outputs or distributions~\cite{cai2025llava,shu2024llava}, implicitly assuming that reproducing answers or static representations suffices to inherit multimodal reasoning ability.

To examine this assumption, we conduct empirical analyses, revealing a pronounced mismatch between textual alignment and visual reasoning:
\textbf{(I)} Correct multimodal reasoning is causally constrained by focused visual attention: when models fail to attend to relevant regions, hallucinated or unreliable responses emerge. 
\textbf{(II)} Even when student models closely match teacher outputs under standard distillation, their visual attention trajectories can diverge substantially, particularly for tasks requiring fine-grained visual grounding.
Together, these findings expose a fundamental \textbf{Perception Gap} in multimodal distillation: students often learn \emph{what to say} without learning \emph{where to look}, instead relying on language priors rather than grounded visual evidence.

Motivated by this insight, we propose \textbf{LaViT}, a distillation framework that aligns latent visual thoughts rather than static visual embeddings. LaViT trains the student to autoregressively generate continuous latent tokens that reconstruct the teacher’s internal visual semantics and attention trajectories prior to textual response generation, explicitly transferring both what visual concepts to encode and where to attend during reasoning.

To prevent shortcut learning through direct access to visual features, we introduce \textbf{Curriculum Sensory Gating}, which progressively restricts and then relaxes visual input during training. This strategy enforces a latent bottleneck early on, compelling reliance on latent visual reasoning while avoiding training--inference mismatch.

Extensive experiments demonstrate that LaViT substantially improves both visual grounding and multimodal reasoning. LaViT-3B achieves up to \textbf{+5.0\%} gains on fine-grained perception benchmarks MMVP~\cite{tong2024eyes} and substantial improvement on BLINK~\cite{fu2024blink}, outperforming strong baselines and rivaling or surpassing 7B models and proprietary GPT-4o.


\section{Related Work}
\subsection{Visual Chain-of-Thought}
Originating from text-only LLMs~\cite{wei2022chain}, Chain-of-Thought (CoT) has expanded to multimodal contexts~\cite{shao2024visual}. Following DeepSeek-R1~\cite{guo2025deepseek}, recent works enhance multi-step visual reasoning via RL-style optimization~\cite{huang2025vision,yang2025r1,shen2025vlm,feng2025video,jiang2025vknowu}; however, these methods primarily rely on indirect textual proxies rather than intrinsic visual understanding. 
Conversely, a parallel stream shifts to \textit{thinking with images} by orchestrating tools~\cite{zhang2025chain,wu2025vtool,su2025openthinkimg,wang2025pixel,zhang2025thyme}, utilizing executable programs~\cite{hu2024visual} or iterative region grounding~\cite{zheng2025deepeyes}. 
Furthermore, unified architectures now support interleaved generation, exemplified by Chameleon's unified tokens~\cite{team2024chameleon}, MVoT's multimodal trajectories~\cite{li2025imagine}, and other general-purpose frameworks~\cite{tong2025metamorph,deng2025emerging,gu2025thinkmorph,cao2026clueaegisheuristictoreasoningcognitiveskilllearning,cao2026revealreasoningenhancedforensicevidence}
\subsection{Latent Reasoning}
Recent advances have shifted the reasoning paradigm from discrete token sequences to continuous hidden states, effectively enhancing both computational efficiency and flexibility~\cite{hao2024training,shen2025codi,wei2025sim}. Extending this concept to multimodal learning, current MLLMs align specialized latent tokens with visual embeddings derived from auxiliary supervision signals, such as helper images~\cite{yang2025machine} or annotated bounding boxes~\cite{li2025latent,chi2026driverwmdrivercentrictrafficconditionedlatent}. To further improve grounding, CoVT~\cite{qin2025chain} integrates fine-grained perceptual priors from models like DINO~\cite{oquab2023dinov2} and SAM~\cite{kirillov2023segment}, while other approaches explore interleaved patterns to mimic internal visual imagination~\cite{tong2025sketch,wang2025monet}. However, these existing methods primarily constrain latent tokens using static encoder features, critically overlooking the dynamic guidance offered by attention maps.

\subsection{Knowledge Distillation}

Knowledge distillation~\cite{hinton2015distilling}, which transfers capabilities from a high-capacity teacher to a compact student, has been widely adopted in LLMs via logit matching~\cite{sun2019patient,jiao2020tinybert}. Extending this paradigm to multimodal models, DistillVLM~\cite{fang2021compressing} performs transformer distillation by using an MSE loss to match the teacher and student’s hidden attention distributions and feature maps. In contrast, MAD~\cite{wang2022multimodal} emphasizes aligning visual and textual token features between teacher and student, leveraging token selection to guide the matching. More recent research explores distillation tailored to MLLMs for specific downstream tasks, including visual grounding~\cite{cai2025llava,feng2025align} and compositional learning~\cite{kim2025compodistill}. 

\section{Empirical Analysis of Perception Gap}
\label{sec:motivation}

We conduct a pilot study to quantify the misalignment between textual generation and visual attention, centered on two research questions: \textbf{(RQ1)} Is correct visual reasoning causally linked to focused visual attention? (i.e., \textit{Does looking at the right place precondition the right answer?}) \textbf{(RQ2)} Does a significant ``alignment gap'' exist in the visual trajectories between teacher and student models, even when their textual outputs are similar?

\subsection{Visual Attention Dictates Reasoning Bounds}
\label{subsec:visual_bounds}
\textbf{Definition 3.1 (Visual Focusing Score).} Let $I$ be the image and $B_{gt}$ the target bounding box. Given the model's aggregated attention trajectory $\mathcal{A}_{traj} \in \mathbb{R}^{H \times W}$, which accumulates attention weights across all layers and heads, the visual focusing score $S_{focus}$ is defined as:
\begin{equation}
    S_{focus} = \frac{\sum_{(u,v) \in B_{gt}} \mathcal{A}_{traj}(u,v)}{\sum_{(u,v) \in I} \mathcal{A}_{traj}(u,v)}
\end{equation}
where $\mathcal{A}_{traj}(u,v)$ denotes the attention intensity at spatial coordinate $(u,v)$. The denominator represents the total attention mass distributed across the entire image $I$. A larger $S_{focus}$ indicates a stronger dependency of the reasoning process on the verified visual evidence, implying that the model is actively ``looking'' at the semantically correct region rather than relying on language priors or blind guessing.

\begin{figure}[t]
  \centering
  \begin{minipage}{0.48\linewidth}
    \centering
    \includegraphics[width=1\linewidth]{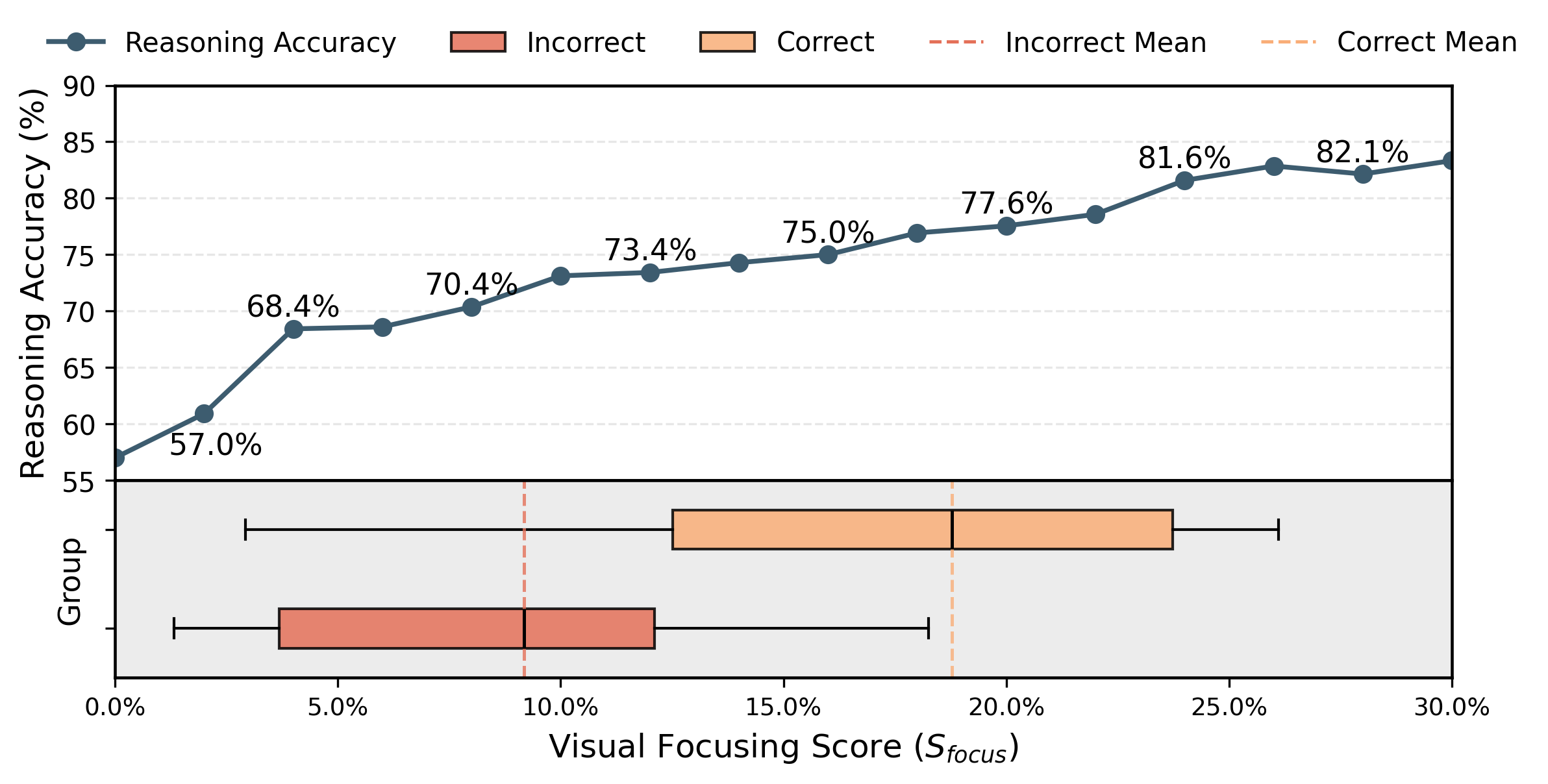} 
    \caption{\textbf{Impact of Visual Attention on Reasoning Accuracy.} The monotonic increase in accuracy with higher Visual Focusing Score ($S_{focus}$) thresholds suggests that effective visual grounding is a critical foundation for correct reasoning.} 
    \label{fig:accuracy_curve}
  \end{minipage}
  \hfill 
  \begin{minipage}{0.48\linewidth}
  \centering
    \includegraphics[width=1\linewidth]{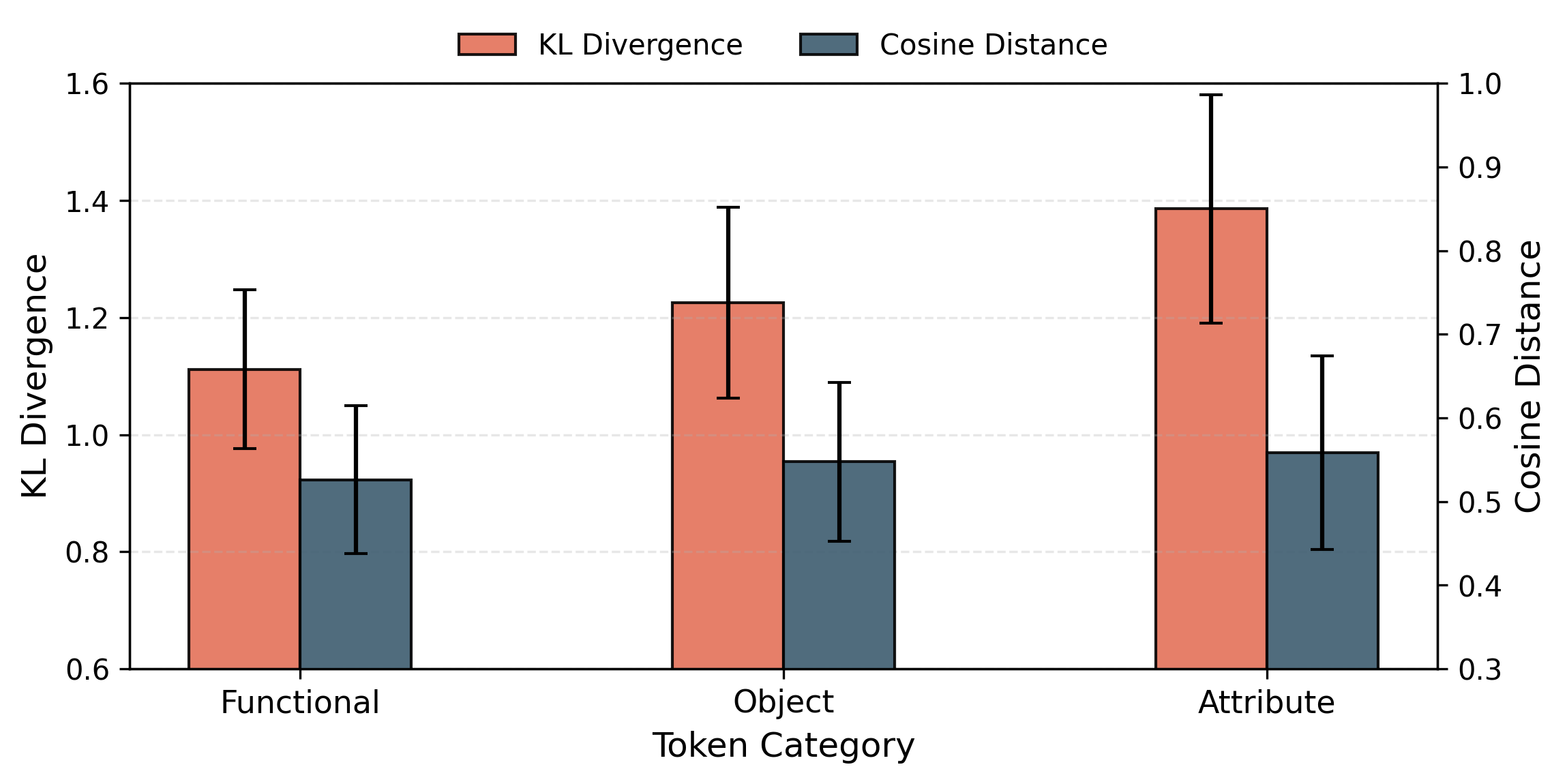}
    \caption{\textbf{The Perception-Reasoning Gap.} While the student aligns closely with the teacher in textual representations (stable Cosine Distance), their visual attention trajectories diverge significantly on attribute-heavy tokens (rising KL Divergence).}
    \label{fig:attention_gap}
    
  \end{minipage}
  \vspace{-2mm} 
\end{figure}

\vspace{0.5em}
\noindent Building upon the above metric, we analyze the attention trajectories of Qwen2.5-VL-32B~\cite{Qwen2.5} on 1,000 randomly sampled instances from the Visual-CoT~\cite{shao2024visual}. As illustrated in Figure~\ref{fig:accuracy_curve}, reasoning outcomes are strictly constrained by the intensity of visual attention:

\begin{itemize}
    \item \textbf{Monotonic Performance Gain:} We observe that reasoning accuracy improves monotonically as the $S_{focus}$ threshold increases. Statistical analysis confirms that correct samples maintain a significantly higher average $S_{focus}$ (\textbf{18.78\%}) compared to incorrect ones (\textbf{9.18\%}), indicating a substantial relative gap of $\sim$104\%. This validates that higher visual energy is a strong predictor of reasoning success.
    \item \textbf{Visual Absence and Hallucination:} Conversely, in samples with negligible $S_{focus}$ ($<1\%$), we predominantly observe responses that are completely irrelevant to the visual content or contain severe hallucinations. This pattern suggests that without active visual grounding, the model relies on language priors to blindly guess, which proves to be a highly unreliable strategy for complex visual tasks.
\end{itemize}

\noindent \fbox{
    \parbox{0.95\linewidth}{
        \textbf{Observation 1:} \textit{\textbf{Visual attention is determinative, not merely interpretative.}} The strict positive correlation confirms that focused visual grounding ($S_{focus}$) is a necessary condition for reasoning success. Models can hardly reason correctly without ``looking'' at the right evidence, effectively ruling out blind guessing as a viable strategy.
    }
}

\subsection{Perception Gap between Teachers and Students}
\label{subsec:perception_gap}

To investigate whether standard Supervised Fine-Tuning (SFT) enables student models to inherit the teacher’s intrinsic visual thinking process, we conduct a controlled experiment using a Naive-SFT baseline trained on the textual components of our proposed LaViT-15k dataset (detailed in Section \ref{method}). Specifically, the student model is fine-tuned to predict the teacher's ground-truth responses using standard cross-entropy loss, without any explicit visual alignment constraints. We then analyze the resulting attention dynamics on a hold-out set of 1,000 samples following the experimental setup in Section \ref{subsec:visual_bounds}.

We fed identical reasoning prompts to both the Teacher and the Student. We categorized the generated tokens into three groups based on their semantic reliance on visual evidence: \textbf{Functional} (e.g., stop words), \textbf{Object} (nouns), and \textbf{Attribute} (adjectives, spatial relations). We then computed the Kullback-Leibler (KL) divergence between their normalized attention maps, alongside the Cosine Distance of their hidden states.

As shown in Figure~\ref{fig:attention_gap}, we observe a decoupling between textual alignment and visual attention:

\begin{itemize}
    \item \textbf{Attention Drift on Visual Concepts:} The attention divergence exhibits a distinct monotonic increase as the semantic reliance on vision deepens. While Functional tokens maintain a lower average divergence ($\mu=1.11$), Attribute tokens, which require precise visual grounding, show the highest misalignment ($\mu \approx 1.39$). This indicates that when describing fine-grained details (e.g., color, texture), the student struggles to focus on the same regions as the teacher.
    \item \textbf{The ``Blind Guessing'' Phenomenon:} Crucially, while the attention divergence surges, the \textit{Cosine Distance} of the hidden states remains relatively stable across categories (ranging from $0.52$ to $0.55$). This implies that the student can mimic the teacher's textual representations (learning \textit{what} to say) without correctly aligning its visual attention (learning \textit{where} to look), effectively relying on language priors rather than active observation.
\end{itemize}

\noindent \fbox{
    \parbox{0.95\linewidth}{
        \textbf{Observation 2:} \textit{\textbf{Textual mimicry does not guarantee visual understanding.}} SFT trains the student to reproduce the teacher's \textit{words} but fails to transfer the underlying \textit{visual trajectory}. This ``Perception Gap'' suggests that the student model is often ``guessing'' based on language context rather than actively ``observing'' the image.
    }
}

\section{Method}
\label{method}

\subsection{LaViT-SFT-15K}
We construct \textbf{LaViT-SFT-15K}, comprising 15K tuples $\langle I, Q, A, \mathcal{A}_{traj}, \mathcal{V}_{sem} \rangle$, to distill the \textbf{Internal Cognitive States} of the teacher (Qwen2.5-VL-32B~\cite{Qwen2.5}). Unlike tool-based approaches, we directly extract intrinsic reasoning traces. Data quality is enforced via a Three-Stage Filtering pipeline: (1) \textbf{Correctness:} retaining samples matching ground truth; (2) \textbf{Difficulty:} removing instances solvable by a text-only model; and (3) \textbf{Alignment:} rejecting samples with $<20\%$ aggregated attention mass falling within the target regions delineated by the ground-truth bounding box annotations in Visual-CoT~\cite{shao2024visual}, thereby excluding non-visually grounded hallucinations.

We extract two white-box signals. First, \textbf{Dynamic Visual Gaze} ($\mathcal{A}_{traj}$) represents the attention trajectory.
Let $Y = \{y_1, y_2, \dots, y_T\}$ denote the response sequence generated by the teacher model given the input image $I$ and instruction $Q$. To capture the intrinsic visual reasoning process, we aggregate the cross-attention weights $W^{(l,h)} \in \mathbb{R}^{T \times P}$ between the generated text tokens in $Y$ and the visual patches $P$. Specifically, for each visual patch $j$, we compute the raw saliency score $S_j$ by averaging the attention weights across all layers $L$, heads $H$, and sequence length $T$:

\begin{equation}
S_j = \frac{1}{L \cdot H \cdot T} \sum_{l=1}^{L} \sum_{h=1}^{H} \sum_{t=1}^{T} W^{(l,h)}_{t,j}
\end{equation}

where $W^{(l,h)}_{t,j}$ represents the attention weight of the $t$-th token in the response sequence $Y$ attending to the $j$-th visual patch in layer $l$ and head $h$. 

We further apply Min-Max normalization to obtain the final gaze probability:
\begin{equation}
    \mathcal{A}_{traj}(j) = \frac{S_j - \min(S)}{\max(S) - \min(S) + \epsilon}
\end{equation}
Crucially, unlike static visual features extracted from a frozen vision encoder, our target $\mathcal{V}_{sem}$ is derived from the Teacher's last transformer layer. Due to the self-attention mechanism, these image token representations have effectively interacted with the textual instructions $Q$. Therefore, $\mathcal{V}_{sem}$ represents contextualized visual thoughts—reflecting not just what is in the image, but how the Teacher interprets the visual content specifically in response to the given query. To distill the most salient visual cues, we subject $\mathcal{A}_{traj}$ to Top-N ($N=8$) sparsification, thereby ensuring sparse and noise-free supervision.

\subsection{White-box Trajectory Distillation}
\label{sec:method_distillation}

We establish the foundational reasoning capability of our model through a Supervised Fine-tuning (SFT) stage defined as \textbf{Latent Teacher Forcing}. Unlike standard SFT that maps inputs directly to text, we train the model to \textbf{autoregressively generate} a sequence of $K$ continuous latent tokens, $\mathbf{V} = \{<v-trace_{1}>, \dots, <v-trace_{k}>\}$, prior to producing the textual response.

These latent tokens serve as \textbf{Visual Information Containers}. By leveraging a white-box distillation approach, we force $\mathbf{V}$ to explicitly capture and compress the teacher's high-dimensional visual semantics and gaze patterns. Formally, given an input $[I, Q]$, the model generates the latent and textual sequences to form the complete trajectory $X = [I, Q, \mathbf{V}, A]$, where $\mathbf{V}$ acts as the indispensable cognitive bridge supplying visual evidence for the subsequent response $A$.

\subsubsection{Curriculum Sensory Gating}
\label{sec:gating}

A naive attention mechanism allows response tokens $A$ to attend directly to image patches $I$, enabling the model to bypass $\mathbf{V}$ (shortcut learning). Conversely, a permanent hard mask creates a training-inference distribution shift. To resolve this, we propose Curriculum Sensory Gating, which modulates the direct visual perception path via a time-dependent scalar $\gamma(t) \in [\epsilon, 1]$.

We implement gating within the attention bias of the Transformer. Let $Q_{txt}$ denote queries from response tokens and $K_{img}$ denote keys from image patches. The attention scores are computed as:
\begin{equation}
\begin{split}
    \text{Attn}(Q_{txt}, K_{img}) &= \text{Softmax}\left( \frac{Q_{txt} K_{img}^\top}{\sqrt{d}} + \mathbf{B}_{gate}(t) \right), \\
    \text{s.t. } \mathbf{B}_{gate}(t) &= \ln(\gamma(t)).
\end{split}
\end{equation}

To ensure a structured internalization process mentioned above, we introduce a warm-up period $T_{w}$. The gating scalar $\gamma(t)$ is defined as:
\begin{equation}
    \gamma(t) = 
    \begin{cases} 
    \epsilon + \frac{1 - \epsilon}{2} \left[ 1 - \cos\left( \frac{\pi t}{T_{w}} \right) \right], & t < T_{w} \\
    1, & t \ge T_{w}
    \end{cases}
\end{equation}
where $T_{w}$ denotes the warm-up steps. This schedule defines two distinct operational phases governed by the training progress:

\begin{itemize}
    \item \textbf{Phase 1: Sensory Warm-up ($t < T_{w}$).} The direct visual path opens gradually, following the cosine curve. Initially, $\gamma \approx \epsilon$ (set to $1\text{e-}6$.). Setting $\epsilon$ to this value not only yields an extremely large negative bias ($B_{gate} \ll 0$) to block the direct visual path---effectively forming an implicit bottleneck---but also maintains the validity of the logarithmic operation, ensuring absolute convergence stability in the early training phase. This creates a strict \textbf{Latent Bottleneck}, mathematically compelling the model to compress necessary visual information into $\mathbf{V}$. The subsequent smooth relaxation prevents optimization shock while establishing a strong dependency on latent reasoning.
    \item \textbf{Phase 2: Fully Observable ($t \ge T_{w}$).} The gate becomes fully open ($\gamma=1$), reducing the bias $\mathbf{B}_{gate}$ to zero. The direct visual path functions as a \textbf{Residual Perception} connection, allowing the model to attend to fine-grained pixel details that complement the high-level reasoning encoded in $\mathbf{V}$. This configuration matches the standard inference topology, ensuring zero distribution shift.
\end{itemize}

\subsubsection{Optimization Objectives}

To ensure the student strictly internalizes the teacher's cognition, we employ a dual-stream distillation scheme with explicit gradient flow controls.

\vspace{1mm}
\noindent \textbf{1. Semantic Reconstruction ($\mathcal{L}_{concept}$).} \\
We align the student's latent hidden states $h_{z}$ with the teacher's holistic visual concepts $V_{\text{sem}}$. Since the teacher's representations encapsulate high-quality visual semantics, we treat them as \textit{fixed semantic anchors}. We employ a projection head $\phi_{\text{mlp}}$ to map the student's latent manifold into the teacher's semantic space:
\begin{equation}
    \mathcal{L}_{concept} = 1 - \frac{1}{B} \sum_{i=1}^{B} \text{CosSim}\left(\phi_{\text{mlp}}(h_{z}^{(i)}), V_{\text{sem}}^{(i)}\right)
\end{equation}
This objective compels the student's latent tokens $\mathbf{V}$ to act as informative containers, actively capturing and compressing the visual information necessary to reconstruct the teacher's visual semantics.

\vspace{1mm}
\noindent \textbf{2. Trajectory Alignment ($\mathcal{L}_{traj}$).} \\
We define the reasoning trajectory as the distribution of attention weights over visual patches. We treat the teacher's attention map $A_{\text{traj}}$ as the target, following prior observations that distilling teacher attention maps can effectively guide student visual alignment~\cite{fang2021compressing}. We constrain the student's attention $A_{\text{student}}$ (originating from $\mathbf{V}$) to match this target via KL Divergence:
\begin{equation}
    \mathcal{L}_{traj} = \frac{1}{B} \sum_{i=1}^{B} \sum_{j=1}^{K} \mathcal{D}_{\text{KL}}\left(A_{\text{traj}}^{(i,j)} \| A_{\text{student}}^{(i,j)}\right).
\end{equation}
This ensures the latent tokens learn where to look, inheriting the teacher's visual search strategy. Crucially, we apply this supervision only to the final latent token $V_{trace\_k}$. This constraint forces the preceding $k-1$ tokens to act as a compression bottleneck, aggregating and refining visual information into a single, cohesive representation before alignment.

\vspace{1mm}
\noindent \textbf{3. Response Generation with Dynamic Gradient Transition ($\mathcal{L}_{ntp}$).} \\
The Next-Token Prediction loss drives the response generation. Crucially, the gradient flow from this loss is intrinsically modulated by our gating mechanism. By chain rule, the sensitivity of the loss with respect to direct visual features is proportional to the attention weight:
\begin{equation}
    \left\| \frac{\partial \mathcal{L}_{ntp}}{\partial I} \right\| \propto \text{Attn}(Q_{txt}, K_{img}) \approx \gamma(t).
\end{equation}
During Phase 1, gradients are fully channeled through $\mathbf{V}$, establishing the bottleneck. As $\gamma \to 1$, the gradients transition to a synergistic flow, optimizing both the latent abstraction and the residual perception paths jointly.

\subsubsection{Joint Training Dynamics}

Unlike complex multi-stage pipelines~\cite{li2025latent,wang2025monet} that require careful hyperparameter scheduling to avoid collapse, our Curriculum Sensory Gating provides a robust structural constraint. This physically enforced bottleneck naturally regulates the learning difficulty, eliminating the need for dynamic loss weighting.

We employ a streamlined \textbf{Joint Training} paradigm with a fixed distillation weight. The total loss is defined as:
\begin{equation}
    \mathcal{L}_{total} = \mathcal{L}_{ntp} + \lambda \cdot (\mathcal{L}_{concept} + \mathcal{L}_{traj}),
\end{equation}
By maintaining a constant but moderate alignment pressure, we ensure the latent tokens $\mathbf{V}$ remain semantically consistent with the teacher in the curriculum, while allowing the NTP loss to primarily drive the generation quality as the sensory gate opens.

\section{Experiment}
\label{exp}

\subsection{Experiment Setup}
\noindent \textbf{SFT Settings.} In Phase 1, we train the model for 400 steps with the sensory gating scalar $\gamma$ initialized at $1\text{e-}6$. In Phase 2, we continue training for an additional 600 steps. We set $\lambda = 0.3$ across all experiments. The model is initialized from Qwen2.5-VL-3B and finetuned on the LaViT-SFT-15K. We set the number of latent visual tokens $V$ to 4. More details are shown in the appendix.

\indent \textbf{Evaluated Benchmarks.} 
We evaluate LaViT on diverse benchmarks. For subtle visual details, we use MMVP~\cite{tong2024eyes} and the Attribute Recognition subset of V$^*$~\cite{wu2024v}, which target CLIP-blind patterns and object attributes. For higher-level cognition, we adopt BLINK~\cite{fu2024blink} tasks on Relative Depth, IQ-Test, Relative Reflectance, and Spatial Relation, which require mental manipulation and geometric reasoning. We also include MMStar~\cite{chen2024we} to test robustness against language priors and ensure genuine visual understanding.

\subsection{Main Results}

\begin{table*}[t]
    \centering
    \begin{minipage}[t]{0.82\textwidth}
        \vspace{0pt} 
        \resizebox{\linewidth}{!}{
            \begin{tabular}{l@{\hspace{3pt}}c cc cccc c}
                \toprule
                & &
                \multicolumn{2}{c}{\cellcolor{BgPerception}\textbf{Fine-grained Perception}} & 
                \multicolumn{4}{c}{\cellcolor{BgReasoning}\textbf{Visual Reasoning}} & 
                \multicolumn{1}{c}{\cellcolor{BgRobustness}\textbf{Robustness}} \\
                \cmidrule(lr){3-4} \cmidrule(lr){5-8} \cmidrule(lr){9-9}
                \textbf{Method} & \textbf{Params} & 
                \textbf{MMVP}\pad &        
                \textbf{V* Attrib.} &    
                \textbf{Rel. Depth}\pad &    
                \textbf{IQ-Test}\pad &    
                \textbf{Rel. Ref.} &      
                \textbf{Spatial}\pad &    
                \textbf{MMStar}\pad \\    
                \midrule
                
                \multicolumn{9}{c}{\cellcolor{HeaderGray}\textit{\textbf{Proprietary Model}}} \\
                GPT-4o~\cite{hurst2024gpt} & - & 58.33 & 72.5 & 64.52 & 30.0 & 38.8 & 76.9 & \underline{63.9} \\
                \midrule
                
                \multicolumn{9}{c}{\cellcolor{HeaderGray}\textit{\textbf{Open Source Model}}} \\
                Qwen2.5-VL~\cite{Qwen2.5} & 7B & 66.7 & 77.39 & 71.77 & 26.0 & 38.8 & 87.4 & 58.9 \\
                Qwen2.5-VL~\cite{Qwen2.5} & 32B & \textbf{75.33} & \underline{82.61} & \underline{75.81} & \underline{30.0} & \textbf{55.97} & 85.31 & \textbf{69.8} \\
                \midrule
                
                \multicolumn{9}{c}{\cellcolor{HeaderGray}\textit{\textbf{Open Source Reasoning MLLMs}}} \\
                Naive SFT & 3B & 65.33 & 80.87 & 70.16 & 28.0 & 34.33 & 81.82 & 55.53 \\
                PAPO~\cite{wang2025perception} & 3B & 50.0* & 22.61* & 59.68 & 31.33* & 33.6 & 76.92 & 52.7 \\
                DMLR~\cite{liu2025reasoning} & 3B & 61.33 & 46.96 & 54.84 & 26.0 & 23.88 & 72.03 & 51.2 \\
                LVR\_RL~\cite{li2025latent} & 3B & 55.3* & 69.6* & 64.52 & 30.7* & 42.54 & 77.62 & 53.73 \\
                R1-OneVision~\cite{huang2025vision} & 7B & 67.0* & - & - & - & - & - & 52.1* \\
                LVR~\cite{li2025latent} & 7B & \underline{72.0} & \textbf{84.4} & 76.61 & 28.7 & 42.5 & \textbf{89.5} & 59.4 \\
                \midrule
                
                \multicolumn{9}{c}{\cellcolor{HeaderGray}\textit{\textbf{Baseline \& Our Model}}} \\
                
                Qwen2.5-VL (Baseline) & 3B & 62.33 & 81.74 & 61.29 & 24.0 & 29.85 & 81.12 & 50.2* \\
                
                \rowcolor{OursHighlight} 
                \textbf{LaViT} & 3B & 
                \res{67.33}{5.00} & 
                \res{\underline{82.61}}{0.87} & 
                \res{\textbf{78.23}}{16.94} & 
                \res{\textbf{32.0}}{8.00} & 
                \res{\underline{45.52}}{15.67} & 
                \res{81.82}{0.70} & 
                \res{54.07}{3.87} \\ 
                \bottomrule
\end{tabular}
        }
        
        \caption{Performance comparison on multimodal benchmarks across three categories: 
        \protect\colorbox{BgPerception}{Fine-grained Perception}, 
        \protect\colorbox{BgReasoning}{Visual Reasoning}, and 
        \protect\colorbox{BgRobustness}{Multimodal Robustness}. 
        The best and second-best results are marked in \textbf{bold} and \underline{underlined}, respectively. 
        Green values indicate the absolute gains of \textbf{LaViT} over the Qwen2.5-VL-3B baseline.
        Asterisks (*) denote results reported by papers.\cite{fu2024blink}\cite{li2025latent}\cite{liu2025reasoning}}
        \label{tab:main_results}
    \end{minipage}%
    \hfill 
    \begin{minipage}[t]{0.17\textwidth}
        \vspace{0pt} 
        \centering
        
        \includegraphics[width=\linewidth, keepaspectratio]{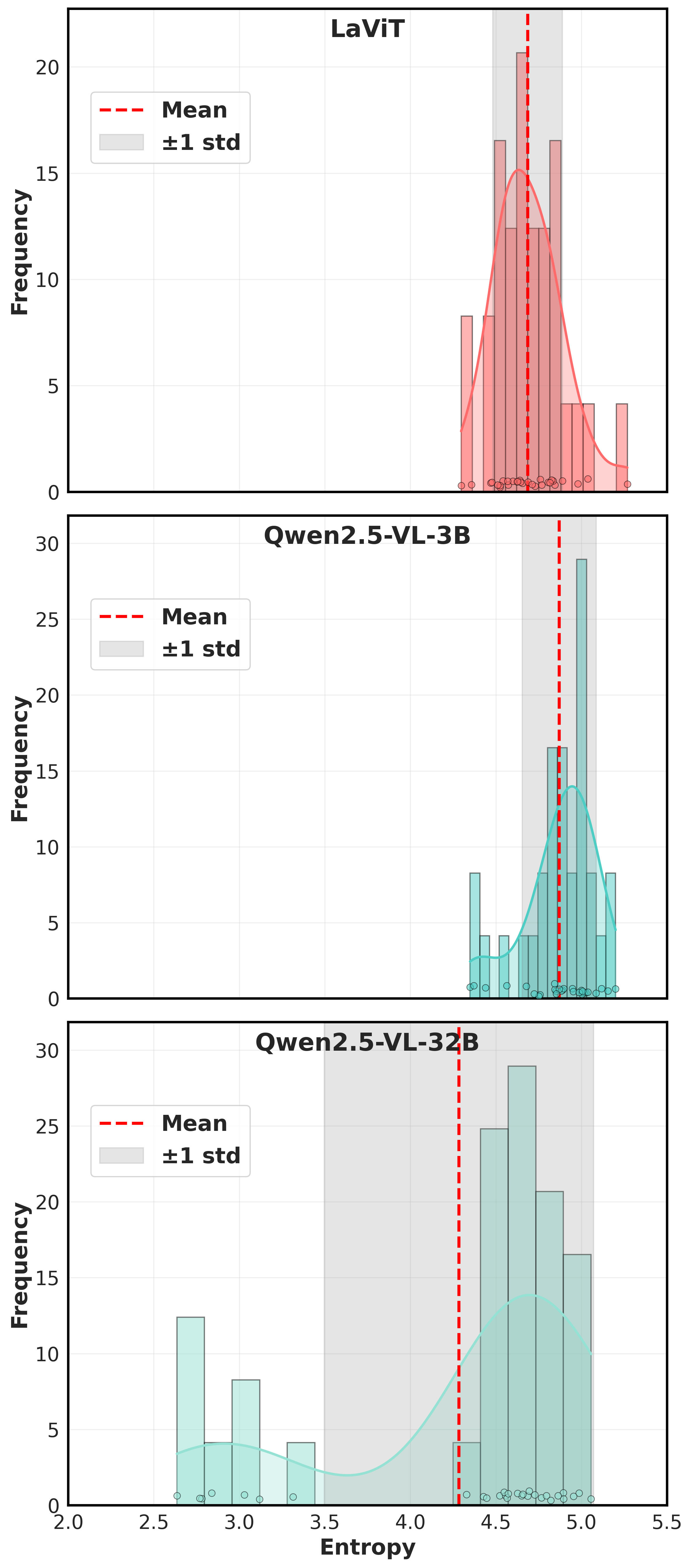}
        
        \captionof{figure}{Attention entropy distribution.\label{fig:side_image}}
        
    \end{minipage}
    \vspace{-5mm}
\end{table*}

\label{sec:main_results}

Table~\ref{tab:main_results} presents the comparative performance of LaViT against state-of-the-art MLLMs. As shown, our method achieves consistent improvements across all benchmarks, demonstrating the efficacy of latent visual thinking.

\noindent \textbf{Cross-Scale Superiority and Efficiency.} 
LaViT significantly enhances its backbone, achieving substantial gains of +15.67\% on Relative Reflectance and +16.94\% on Relative Depth. Despite its compact 3B scale, it demonstrates remarkable cross-scale competitiveness, outperforming the larger Qwen2.5-VL-7B on five out of seven benchmarks. Moreover, LaViT surpasses SOTA 7B models, beating LVR-7B on fine-grained spatial tasks (e.g., Relative Depth: 78.23\% vs. 76.61\%) and R1-OneVision-7B on MMStar. These results confirm that optimizing latent thinking is a more parameter-efficient strategy than simply scaling up model size.

\noindent \textbf{Advantage in Complex Visual Reasoning.} 
LaViT excels in perception-intensive BLINK benchmarks by leveraging continuous latent reasoning to preserve spatial structures, effectively addressing the limitations of standard models in abstract geometric manipulation. Consequently, LaViT-3B achieves 78.23\% on Relative Depth and 32.0\% on IQ-Test, outperforming proprietary models like GPT-4o (64.52\% and 30.0\%) and reasoning-enhanced baselines. Notably, it even surpasses the SOTA latent reasoning model LVR-7B on Relative Depth (+1.62\%) and Relative Reflectance (+3.0\%), validating its superior capability in capturing structural visual semantics.

\noindent \textbf{Fine-Grained Perception and Robustness.} 
LaViT effectively mitigates ``CLIP-blindness,'' achieving 67.33\% on MMVP and substantially outperforming DMLR (61.33\%) and PAPO (50.0\%). By refining visual features to correct encoding errors rather than trading perception for reasoning, LaViT ensures robust visual grounding. This is further evidenced by its 54.07\% score on MMStar (vs. 50.2\% baseline), confirming that performance gains stem from genuine visual understanding rather than language hallucinations.

\subsection{In-depth Analysis of Attention Dynamics}
\label{sec:analysis}

To investigate the underlying mechanisms of LaViT, we conduct a deep dive into the attention distribution on the BLINK \textit{Relative Depth}. We combine quantitative metrics with qualitative visualizations to analyze \textbf{Concentration} and \textbf{Stability}.

\begin{figure*}[t]  
    \centering
    \vspace{-3mm}
    \includegraphics[width=\textwidth]{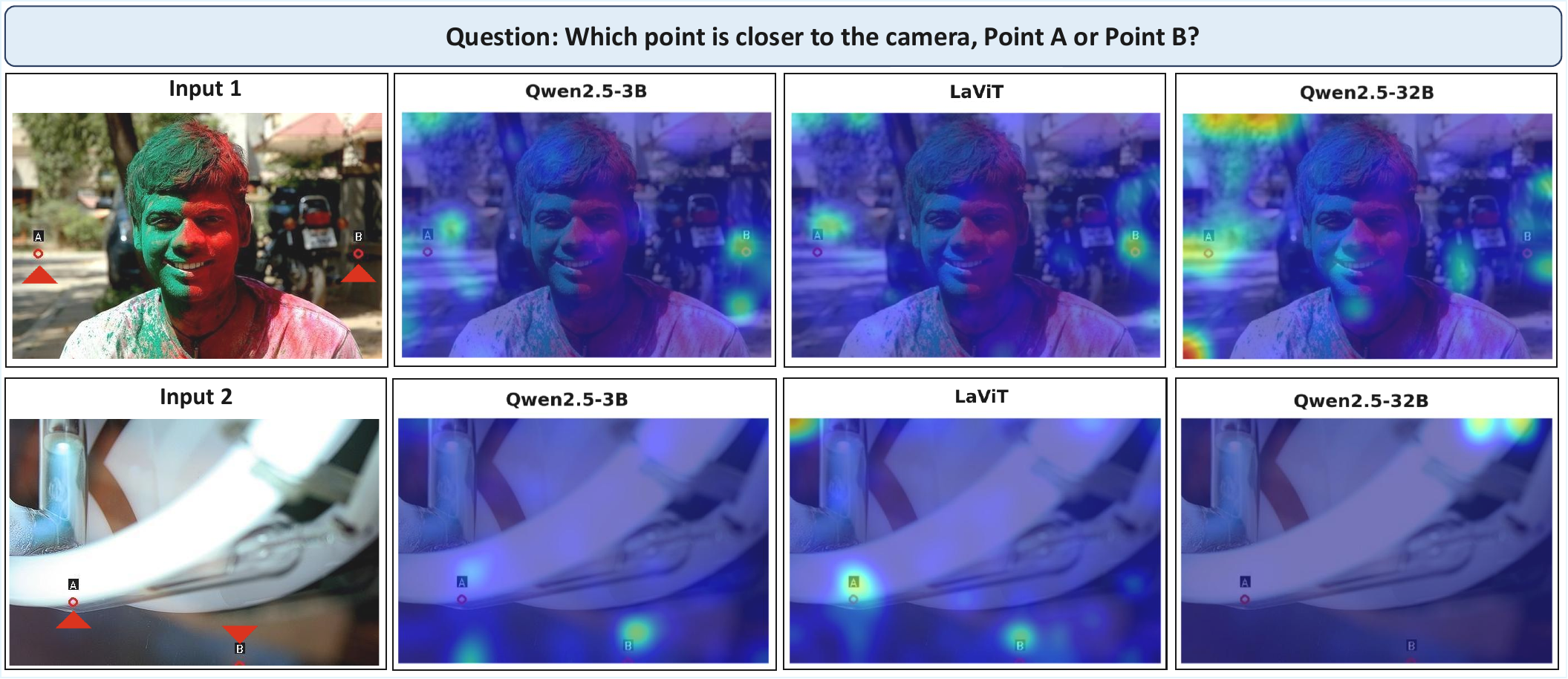}
    \vspace{-7mm}
    \caption{Visualization of attention distributions across Qwen2.5-VL-3B, 32B (Teacher), and LaViT on two representative samples from the BLINK. \textcolor{red}{$\blacktriangle$} indicates the task-relevant critical regions required for correct reasoning.}
    \vspace{-3mm}
\label{fig:attention_visualization}
\end{figure*}

\paragraph{Quantifying Attention Concentration (Entropy).}
We employ Information Entropy ($H$) to measure the ``sharpness'' of the model's visual focus. Formally, for a given image, the attention entropy is defined as:
\begin{equation}
    H = -\sum_{i=1}^{N} p_i \log(p_i)
    \label{eq:entropy}
\end{equation}
where $p_i$ is the normalized attention weight of the $i$-th patch. To quantify these observations, we analyze the attention statistics on the BLINK Relative Depth dataset (Table 2). We define Salient Regions as patches with attention weights surpassing $1.5 \times$ the global mean, serving as a proxy for the area of active visual processing. Additionally, we report the Coefficient of Variation (CV) ($\sigma/\mu$) to measure the stability of the attention mechanism. As visualized in Figure~\ref{fig:side_image} and detailed in Table~\ref{tab:attention_stats}, the Base 3B model exhibits a heavy tail towards high entropy ($H=4.870$), suggesting it lacks a clear visual target. LaViT significantly shifts this distribution leftward, reducing the mean entropy to \textbf{4.686}. This improvement not only approximates the Teacher's focused state ($H=4.284$) but is also visually corroborated by Figure~\ref{fig:attention_visualization}, where LaViT exhibits pinpoint focus on critical depth markers compared to the scattered gaze of the Base model.

\begin{tcolorbox}[colback=gray!10,colframe=gray!40,arc=2mm,title=\textbf{\textit{Takeaway 1: Sharpening Visual Focus}}]
LaViT effectively transitions the student from a ``diffuse'' observation mode to a ``focused'' reasoning mode. By reducing attention entropy (Figure~\ref{fig:side_image}), the student learns to ignore irrelevant background noise and concentrate on task-relevant regions.
\end{tcolorbox}

\begin{table}[t]
    \centering
    \begin{minipage}[t]{0.48\textwidth}
        \centering
        \caption{Statistical analysis of attention distribution on BLINK Relative Depth.}
        \label{tab:attention_stats}
        \resizebox{\linewidth}{!}{%
            \begin{tabular}{l|cccc|ccc}
            \toprule
            \multirow{2}{*}{\textbf{Model}} & \multicolumn{4}{c|}{\textbf{Salient Regions Count}} & \multicolumn{3}{c}{\textbf{Attention Entropy ($H$)}} \\
            \cmidrule(lr){2-5} \cmidrule(lr){6-8}
             & Mean & Std ($\sigma$) & Range & CV & Mean & Std ($\sigma$) & CV \\
            \midrule
            Qwen2.5-32B  & 39.9 & 15.6 & 8-65 & 0.392 & \textbf{4.284} & 0.787 & 0.1837 \\
            Qwen2.5-3B     & 53.8 & 9.0  & 36-77 & 0.191 & 4.870 & 0.216 & 0.0444 \\
            \rowcolor{gray!10} \textbf{LaViT} & 47.3 & \textbf{5.6} & \textbf{43-69} & \textbf{0.102} & 4.686 & \textbf{0.204} & \textbf{0.0435} \\
            \bottomrule
            \end{tabular}%
        }
    \end{minipage}\hfill 
    \begin{minipage}[t]{0.5\textwidth}
        \centering
        \caption{
            Ablation study on key components of LaViT.
        }
        \label{tab:ablation}
        \setlength{\tabcolsep}{1.5pt}
        \resizebox{\linewidth}{!}{%
            \begin{tabular}{lccccc}
            \toprule
            \textbf{Method} & \textbf{MMVP} & \textbf{Rel. Depth} & \textbf{IQ-test} & \textbf{Rel. Ref} & \textbf{Spatial} \\ 
            \midrule
            LaViT-3B & \textbf{67.33} & \textbf{78.23} & \textbf{32} & 45.52 & \textbf{81.82} \\ 
            \hspace{1em} w/o Traj. Align ($L_{traj}$) & 64.33 & 75 & 30.67 & 44.03 & 78.32 \\
            \hspace{1em} w/o Sem. Recon ($L_{concept}$ & 65.33 & 75.81 & 30.67 & 42.54 & 76.92 \\
            \hspace{1em} w/o Curr. Gate & 59.33 & 71.77 & 27.33 & \textbf{48.51} & 79.72 \\
            \hspace{1em} w/o Latent Tokens & 64.33 & 77.42 & 25.33 & 81.12 & 25.33 \\ 
            \hspace{1em} Single Stage Training & 65.67 & 71.77 & 28 & 38.81 & 79.02 \\ 
            \bottomrule
            \end{tabular}%
        }
    \end{minipage}
\end{table}

\paragraph{Achieving Superior Stability via Sparsification.}
While the Teacher model is highly focused, it exhibits significant variance in its viewing strategy across samples (Salient Regions CV = 0.392). We observe that LaViT not only inherits the Teacher's focus but actually achieves a much more stable attention pattern (CV = \textbf{0.102}), surpassing the teacher in consistency.

We attribute this distilled stability directly to our \textbf{White-box Trajectory Distillation} design:
\begin{itemize}
    \item \textbf{Top-N Sparsification:} By explicitly supervising the student with only the Top-N ($N=8$) strongest attention points from the teacher, we enforce a hard constraint that filters out the teacher's low-confidence ``attentional noise'' or hesitation.
    \item \textbf{Data Filtering:} Our preprocessing pipeline excludes samples where attention does not align with ground-truth regions, ensuring only high-quality traces are learned.
\end{itemize}

Consequently, LaViT does not merely mimic the Teacher's raw output; it distills the \textbf{most robust visual cues}, resulting in a student model that is consistently focused on critical regions without the variance observed in the large teacher model.

\begin{tcolorbox}[colback=gray!10,colframe=gray!40,arc=2mm,title=\textbf{\textit{Takeaway 2: The Denoising Effect}}]
LaViT acts as a \textbf{semantic filter}. Instead of inheriting the teacher's instability, LaViT leverages Top-N sparsification to retain only the core attention patterns, reducing the Coefficient of Variation (CV) from 0.392 to 0.102. This results in a student who is surprisingly more decisive and stable than its teacher.
\end{tcolorbox}

\subsection{Ablation Study}
\label{sec:ablation}

We evaluate LaViT on MMVP and BLINK subsets, comparing it against variants including single-stage training and the removal of curriculum gating.

\noindent\textbf{Impact of Alignment Components.} To dissect the contribution of each module, we report the ablation results in Table~\ref{tab:ablation}. We investigate the impact of removing key components:
\begin{itemize}
    \item \textbf{Trajectory Alignment (w/o Traj.):} Removes the attention alignment loss ($L_{traj}$).
    \item \textbf{Semantic Reconstruction (w/o Sem.):} Excludes the concept reconstruction objective ($L_{concept}$).
    \item \textbf{Curriculum Sensory Gating (w/o Gate):} Replaces our progressive schedule with a hard switch ($\gamma=0$ in Phase 1, $\gamma=1$ in Phase 2).
    \item \textbf{Single Stage Training:} Conducts training in a single phase where visual tokens are always fully visible ($\gamma(t)=1$).
    \item \textbf{Latent Tokens (w/o Latent):} Masks the generated latent visual tokens during inference to test dependency.
\end{itemize}

As shown in Table~\ref{tab:ablation}, removing either Trajectory Alignment or Semantic Reconstruction leads to significant performance drops across all tasks. Furthermore, masking latent tokens during inference precipitates a marked decline, verifying the model's genuine reliance on the generated visual thoughts. This confirms explicit supervision on ``where to look'' and ``what to see'' is essential for the student to inherit the teacher's visual cognitive patterns effectively.

\noindent\textbf{Effectiveness of Curriculum Sensory Gating.} 
The performance degradation observed without this module (e.g., MMVP accuracy drops to 59.33\%) indicates that a simple hard switch of visual visibility is insufficient. 
Our progressive gating strategy is crucial for preventing shortcut learning and forcing the model to rely on deep latent reasoning.

\noindent\textbf{Impact of Training Strategy.} 
The \textit{Single Stage Training} baseline ($\gamma(t)=1$) consistently underperforms the full model, particularly on Relative Reflectance (38.81\% vs. 45.52\%). 
This demonstrates that progressively exposing visual information is key to avoiding sub-optimal convergence and building robust reasoning capabilities.

\section{Conclusion}

In this work, we identify the ``Perception Gap'' in multimodal distillation, where student models often mimic textual outputs without inheriting the teacher's visual attention patterns. To bridge this, we propose \textbf{LaViT}, a framework that aligns latent visual thoughts via White-box Trajectory Distillation and Curriculum Sensory Gating. By utilizing latent tokens as cognitive containers, LaViT compels the student to reconstruct the teacher's visual semantics and gaze before response generation. Extensive experiments demonstrate that LaViT-3B significantly outperforms SFT baselines and rivals 7B-scale models on reasoning-intensive benchmarks like BLINK and MMVP.

\section*{Acknowledgements}
Acknowledgments are omitted for double-blind review.

\clearpage
\appendix
\renewcommand{\theHsection}{appendix.\Alph{section}}
\renewcommand{\theHsubsection}{appendix.\Alph{section}.\arabic{subsection}}
\section{Implementation and Reproducibility Details}
\label{appendix}

This appendix provides supplementary materials, including detailed hyperparameter configurations, further experimental analyses, and comprehensive statistics regarding our dataset construction. To facilitate reproducibility and support the open-source community while strictly adhering to the double-blind review policy, we have released our complete codebase, pre-trained models, and the LaViT-SFT-15k dataset in an anonymous open-source repository available at:\textcolor{blue}{https://anonymous.4open.science/r/LaViT/}

\subsection{Baselines}
To comprehensively evaluate the effectiveness of LaViT, we benchmark it against a diverse spectrum of state-of-the-art MLLMs. We categorize these baselines into three distinct paradigms to isolate the contributions of our latent reasoning mechanism from model scale and data exposure.

\paragraph{General-Purpose Foundation Models.}
We first establish a performance lower bound using our backbone model, \textbf{Qwen2.5-VL-3B}~\cite{Qwen2.5}, to quantify the specific gains attributed to our architecture. Crucially, to prove that our improvements stem from the proposed \textit{latent thinking mechanism} rather than merely domain-specific data exposure, we construct a controlled baseline named \textbf{Naive-SFT}. This model is fine-tuned on the identical LaViT-SFT-15k dataset but utilizes standard text-only supervision, serving as a rigorous control variable. Furthermore, we include \textbf{Qwen2.5-VL-7B} to assess cross-scale competitiveness and employ \textbf{GPT-4o}~\cite{hurst2024gpt} (\textit{gpt-4o-2024-08-06}) as a proprietary upper-bound reference to gauge how close our compact 3B model is to industrial state-of-the-art performance.

\paragraph{Explicit Reasoning Frameworks (Thinking-about-Images).}
This category includes methods that enhance reasoning via explicit textual Chains-of-Thought (CoT) or Reinforcement Learning. We compare against \textbf{R1-OneVision}~\cite{yang2025r1}, which explicitly generates a ``think-before-answer'' trajectory in the language space. Additionally, we include \textbf{PAPO}~\cite{wang2025perception}, an RL-based approach that optimizes image-grounded descriptions through verifiable rewards. For a fair comparison, we reproduce PAPO on the same 3B backbone to evaluate the efficiency of latent versus explicit alignment strategies.

\paragraph{Latent Visual Reasoning Competitors.}
Finally, we compare LaViT against direct competitors that also operate within the latent space. We benchmark against \textbf{LVR}~\cite{li2025latent} (and its RL variant), which operates on a larger 7B backbone. This serves as a cross-scale benchmark to demonstrate LaViT's parameter efficiency. We also compare with \textbf{DMLR}~\cite{liu2025reasoning}, a framework utilizing test-time latent optimization. To ensure a fair comparison regarding computational cost and inference latency, we re-implement DMLR on the Qwen2.5-VL-3B backbone with the number of latent optimization steps restricted to 4, matching the computational budget of our method.

\subsection{Hyperparameters}
\label{app:hyperparameters}

The Table \ref{tab:hyperparams} details the hyperparameters used for the 1000-step training run.
\begin{table*}[t]
    \centering
    \small 
    \caption{Hyperparameters used for the 1000-step training run.}
    \label{tab:hyperparams}
    
    \begin{tabularx}{\textwidth}{l l l X} 
    \toprule
    \textbf{Category} & \textbf{Parameter} & \textbf{Value} & \textbf{Description} \\
    \midrule
    \multirow{5}{*}{Optimization} 
        & \textbf{Learning Rate} & \textbf{5e-6} & Initial learning rate \\
        & LR Scheduler & linear & Linear decay with warmup \\
        & Warmup Ratio & 0.03 & Default transformer warmup \\
        & Optimizer & AdamW & $\beta_1=0.9, \beta_2=0.999$ \\
        & Weight Decay & 0.0 & No weight decay applied \\
    \midrule
    \multirow{3}{*}{Training Scale} 
        & Total Steps & 1000 & Fixed step training \\
        & Batch Size & 16 & Per-device training batch size \\
        & Grad. Accum. & 1 & No gradient accumulation \\
    \midrule
    \multirow{4}{*}{Model Config} 
        & Num Latent Tokens & 4 & Number of latent tokens \\
        & V-top Dim & 5120 & Dimension of reconstruction target \\
        & \textbf{Freeze Vision} & \textbf{True} & ViT encoder weights are locked \\
        & \textbf{Freeze LLM} & \textbf{False} & LLM backbone is fine-tuned \\
    \midrule
    \multirow{2}{*}{Data Config} 
        & Max Pixels & 1,003,520 & 1280$\times$28$\times$28 equivalent \\
        & Min Pixels & 200,704 & 256$\times$28$\times$28 equivalent \\
    \bottomrule
    \end{tabularx}
\end{table*}

\subsection{Computational Cost and Efficiency}
\label{sec:computational_cost}

In this section, we analyze the computational cost incurred by the proposed latent reasoning components. We measure clock-time with batch size 1 on a single NVIDIA RTX Pro 6000 96GB GPU using the Hugging Face \texttt{transformers} library to avoid hardware-level throughput optimizations and accurately reflect architectural latency. We evaluate LaViT alongside the base Qwen2.5-VL-3B, a text-only SFT baseline, DMLR\cite{liu2025reasoning}, and LVR\cite{li2025latent} on the full MMVP\cite{tong2024eyes} dataset (300 samples). 

As shown in Table~\ref{tab:comp_cost}, we report the Time-to-First-Token (TTFT), latent generation time, text generation time per token, and the average total inference seconds. Compared to the base model and the text-only SFT baseline, LaViT introduces a modest latency overhead during the latent generation phase (91.26 ms). However, this overhead is significantly lower than that of other latent reasoning frameworks, such as LVR (156.83 ms) and the test-time optimization method DMLR (956.62 ms). 

Remarkably, LaViT achieves the most efficient text generation speed (15.61 ms/token), substantially outperforming both the base model (28.83 ms/token) and the text-only SFT baseline (42.47 ms/token). This efficiency gain during the autoregressive decoding phase can be attributed to the highly compressed nature of our continuous latent tokens, which effectively reduces the KV-cache burden. Overall, LaViT maintains a highly competitive average total inference time of 0.143 seconds, demonstrating that our latent visual thinking mechanism achieves superior reasoning performance without compromising practical inference efficiency.

\begin{table}[h]
\centering
\small
\caption{Inference efficiency comparison on the full MMVP dataset. Both DMLR and LVR are implemeneted on Qwen2.5-VL-3B}
\label{tab:comp_cost}
\resizebox{\linewidth}{!}{%
\begin{tabular}{lcccc}
\toprule
\textbf{Method} & \textbf{TTFT (ms)} & \textbf{Latent Gen. (ms)} & \textbf{Text Gen. (ms/token)} & \textbf{Total Avg. (s)} $\downarrow$ \\
\midrule
Text-only SFT & 43.75 & - & 42.47 & 0.062 \\
Qwen2.5-VL-3B (Base) & \textbf{36.48} & - & 28.83 & \textbf{0.063} \\
\midrule
\textbf{LaViT (Ours)} & 43.59 & \textbf{91.26} & \textbf{15.61} & 0.143 \\
LVR & 72.44 & 156.83 & 16.90 & 0.215 \\
DMLR & 86.79 & 956.62 & 47.41 & 1.177 \\
\bottomrule
\end{tabular}
}
\end{table}

\subsection{Evaluation Protocols}
\label{sec:eval_protocols}

To ensure standardized, fair, and reproducible comparisons across all baseline models and our proposed method, we conduct all benchmark evaluations using VLMEvalKit~\cite{duan2024vlmevalkit}. This open-source evaluation suite guarantees consistent prompt formatting, inference pipelines, and metric calculations across all the vision-language datasets reported in this work, effectively eliminating potential performance discrepancies arising from varying evaluation implementations.

\newpage
\section{In-depth Ablation Studies}
\subsection{Analysis of numbers of continuous latent tokens $K$}
To investigate the optimal capacity of the latent bottleneck, we conducted an ablation study on the number of latent visual tokens $K \in \{4, 6, 8\}$ and monitored performance variations across different training steps. As detailed in Table \ref{tab:ablation_k}, we observe that $K=4$ yields the superior balance between visual grounding and reasoning capabilities.Specifically, the model with $K=4$ achieves peak performance at 1,000 steps, recording the highest scores on MMVP (67.33) and IQ-Test (32.0), while matching the best Relative Reflectance performance (45.52). Interestingly, increasing the number of latent tokens to $K=6$ or $K=8$ does not translate to performance gains; instead, it leads to a slight degradation in reasoning tasks (e.g., IQ-Test and Relative Reflectance). This suggests that a compact set of 4 latent tokens is sufficient to encapsulate the necessary high-level visual semantics guided by the teacher's attention, whereas a larger $K$ may introduce redundancy or noise into the reasoning process. Furthermore, regarding training dynamics, we observe that performance generally peaks around 1,000 steps before stabilizing or slightly declining, indicating that the model reaches optimal alignment at this stage. Consequently, we adopt $K=4$ and the 1,000-step checkpoint for all main experiments reported in this paper.
\begin{table}[h]
\centering
\caption{Ablation study on the number of latent tokens ($K$) and training steps. We report the performance on MMVP and BLINK subsets (IQ-Test, Relative Reflectance, and Spatial Relation). The selected configuration ($K=4$, 1000 steps) is highlighted in \textbf{bold}.}
\label{tab:ablation_k}
\begin{tabular}{lcccccc}
\toprule
\textbf{Method} & \textbf{$K$} & \textbf{Steps} & \textbf{MMVP} & \textbf{IQ-Test} & \textbf{Rel. Ref.} & \textbf{Spatial} \\
\midrule
Qwen2.5-VL-3B & - & - & 62.33 & 24.00 & 29.85 & 81.12 \\
Naive SFT & - & - & 65.33 & 28.00 & 34.33 & 81.82 \\
\midrule
\multirow{5}{*}{LaViT} & \multirow{5}{*}{4}  
  & 600 & 58.33 & 26.00 & \textbf{45.52} & 83.92 \\
& & 800 & 63.33 & \textbf{32.00} & 44.78 & 77.62 \\
& & \textbf{1000} & \textbf{67.33} & \textbf{32.00} & \textbf{45.52} & \textbf{81.82} \\
& & 1200 & 67.00 & 28.67 & 42.54 & 76.92 \\
& & 1400 & 67.67 & 26.67 & 43.28 & 78.32 \\
\midrule
\multirow{5}{*}{LaViT} & \multirow{5}{*}{6}  
  & 600 & 62.00 & 26.67 & 42.54 & 80.42 \\
& & 800 & 58.67 & 28.67 & 43.28 & 81.82 \\
& & 1000 & 61.33 & 27.33 & 43.28 & 77.62 \\
& & 1200 & 63.00 & 26.00 & 42.54 & 76.92 \\
& & 1400 & 65.67 & 28.67 & 44.78 & 75.52 \\
\midrule
\multirow{5}{*}{LaViT} & \multirow{5}{*}{8}  
  & 600 & 63.33 & 22.67 & 30.60 & 84.62 \\
& & 800 & 66.00 & 28.00 & 42.54 & 77.62 \\
& & 1000 & 65.67 & 28.00 & 38.81 & 79.02 \\
& & 1200 & 66.00 & 28.67 & 40.30 & 76.92 \\
& & 1400 & 66.67 & 29.33 & 40.30 & 76.22 \\
\bottomrule
\end{tabular}
\end{table}

\subsection{Data Scaling Efficiency}
\label{sec:data_scaling}

We investigate the data scalability of our proposed method by tracking the performance of models trained on varying sizes of data subsets, ranging from 3K to 15K samples. Crucially, to ensure a rigorously controlled evaluation, we keep the total training iterations fixed across all experimental runs. Consequently, models trained on smaller data subsets (e.g., 3K) undergo more training epochs over the same data to reach the iteration target. We compare the standard SFT baseline and LaViT across three diverse evaluation metrics: MMVP (fine-grained perception), Jigsaw, and Relative Depth (spatial reasoning). 

\begin{figure}[htbp]
    \centering
    \includegraphics[width=\linewidth]{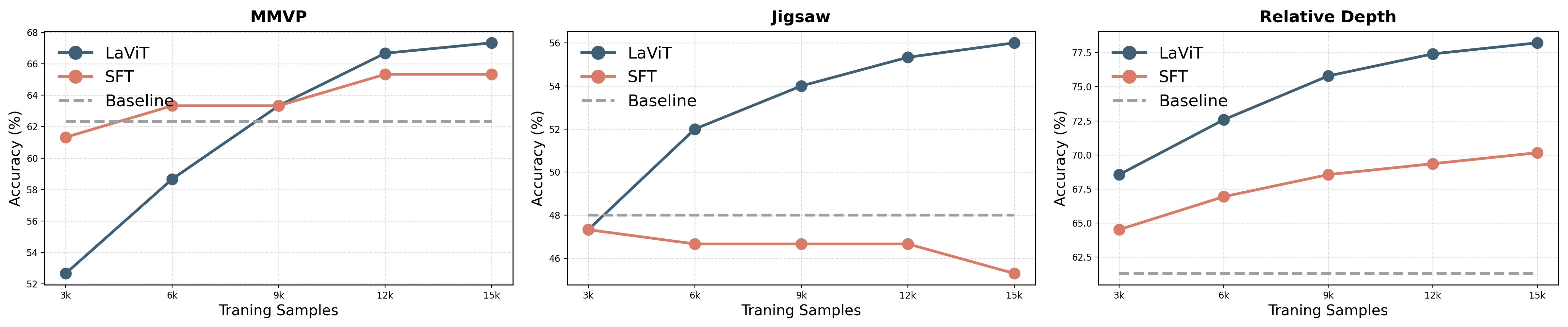} 
    
    \caption{Data scaling efficiency comparison between the standard SFT baseline and LaViT across MMVP, Jigsaw, and Relative Depth tasks from 3k to 15k samples.}
    \label{fig:data_scaling}
\end{figure}

As illustrated in Figure~\ref{fig:data_scaling}, LaViT consistently improves performance across all three benchmarks as the number of unique training samples grows. In stark contrast, the standard SFT baseline quickly saturates (e.g., plateauing at 65.33 on MMVP beyond 12K samples) and even suffers from severe performance degradation on complex spatial tasks. Specifically, under our fixed-iteration setup, repeated exposure to limited visual-text pairs causes the text-only SFT model to severely overfit, driving its Jigsaw score down from the zero-shot baseline's 48.00 to 45.30 when trained on 15K samples. 

Notably, LaViT demonstrates remarkable data scaling efficiency: our model trained on only 9K samples significantly outperforms the standard SFT model trained on the full 15K dataset in both Jigsaw (54.00 vs. 45.30) and Relative Depth (75.80 vs. 70.16). While LaViT initially exhibits lower performance on MMVP when restricted to 3K samples (52.67)—likely because optimizing continuous latent bottleneck representations on a highly limited, homogeneous data pool hinders generalization—it rapidly overcomes this data-scarcity bottleneck as sample diversity increases. By 15K samples, it achieves a peak MMVP performance of 67.33. This compellingly demonstrates that our latent visual thinking mechanism not only effectively mitigates the overfitting issues prevalent in naive SFT but also facilitates the extraction of richer, more generalizable representations, thereby unlocking superior data scalability.

\section{Training Data Construction}
\label{app:data_construction}

\subsection{Data Enrichment}
To support the proposed distillation framework, the training dataset was enriched with pre-computed visual features serving as teacher supervision signals:
\begin{enumerate}
    \item \textbf{V-top Tensors:} High-dimensional feature vectors (5120-dim) extracted from the final layer of the base model's visual encoder, representing holistic semantic concepts.
    \item \textbf{Attention Maps:} Compressed attention weights aggregated across heads and layers, used to generate the trajectory supervision signal for explicit visual grounding.
\end{enumerate}

\subsection{Data Processing and Alignment}
To ensure precise synchronization between the student's latent states and the teacher's signals, we implemented the following processing strategies:

\paragraph{Adaptive Scaling.} 
Given the dynamic resolution characteristics of Qwen2.5-VL, feature maps often vary in spatial dimensions. We record the critical step of applying \textbf{Bilinear Interpolation} to align the spatial resolution of the pre-computed \texttt{v\_top} feature maps with the \texttt{attention\_maps}. This step is essential for maintaining pixel-level correspondence across inputs with varying aspect ratios.

\paragraph{Latent Supervision Strategy.} 
For the optimization of latent visual thoughts, we explicitly designate the hidden state of the \textbf{last latent token} (e.g., \texttt{<v-trace4>}) as the anchor for supervision. By computing the loss only at the end of the latent sequence, we force the visual information to flow fully through the bottleneck, compelling the preceding latent tokens to compress and structure the visual data effectively.

\subsection{System Prompt for Alignment}
During the Supervised Fine-Tuning (SFT) stage, we utilized a specialized system prompt to prime the model for latent visual reasoning. The prompt is presented below:

\begin{center}
\begin{tcolorbox}[
    colback=gray!10,       
    colframe=gray!50,      
    title=\textbf{System Prompt}, 
    arc=1mm,               
    boxrule=0.5pt,         
    width=0.95\linewidth   
]
``You are an expert multimodal large language model. You process visual information through specialized latent tokens to ensure precise alignment between visual perception and textual reasoning.''
\end{tcolorbox}
\end{center}

\subsection{Details of LaViT-SFT-15k Construction}

To ensure the high quality of the distilled internal cognitive states, we implement a rigorous three-stage filtering pipeline.

\paragraph{Correctness Verification} 
For Correctness Verification, we employ MiMo-V2-Flash~\cite{xiao2026mimo} as an LLM-as-a-judge to evaluate the teacher model's generated responses. We prompt the judge with the original instruction, the ground-truth answer, and the generated textual response to determine if the final deduction aligns perfectly with the ground truth. Only strictly correct samples are retained. The system prompt is structured as follows:

\begin{center}
\begin{tcolorbox}[
    colback=gray!10,       
    colframe=gray!50,      
    title=\textbf{System Prompt for Correctness Verification}, 
    arc=1mm,               
    boxrule=0.5pt,         
    width=0.95\linewidth   
]
``You are an expert evaluator tasked with verifying the reasoning correctness of a multimodal large language model. You will be provided with a [Question], a [Ground-Truth Answer], and a [Model-Generated Response]. 

Your task is to determine whether the [Model-Generated Response] arrives at the correct conclusion as defined by the [Ground-Truth Answer]. You must carefully evaluate the logical consistency of the reasoning steps and ensure the final deduction strictly matches the ground truth.

Please provide a binary evaluation:
- Output 'CORRECT' if the model's reasoning is sound and the final answer perfectly aligns with the ground truth.
- Output 'INCORRECT' if the final answer contradicts the ground truth, or if the reasoning contains severe logical flaws and hallucinations.

Provide your evaluation in the format: Decision: [CORRECT/INCORRECT]''
\end{tcolorbox}
\end{center}

\paragraph{Difficulty Control}
To enforce appropriate Difficulty, we filter out instances that can be solved without visual information. Specifically, we input the textual instruction into a text-only model, Gemini-2.5-Flash~\cite{comanici2025gemini25}, completely omitting the accompanying image. If the model successfully predicts the ground-truth answer relying solely on language priors, the sample is deemed too simple and is discarded. This ensures that the remaining instances in LaViT-SFT-15K strictly require genuine multimodal reasoning.

\paragraph{Visual Alignment}
To guarantee precise spatial focus, we perform Visual Alignment using Attention-BBox Overlap Filtering. Utilizing the ground-truth bounding box annotations from Visual-CoT, we extract the teacher model's attention weights over the visual tokens for each reasoning step. These tokens are mapped to the image patch grid to construct a spatial mask corresponding to the target regions. We then calculate the ratio of the attention mass falling inside the bounding boxes to the total visual attention weight. A sample is accepted if this ratio reaches or exceeds a threshold of $\tau = 0.20$ during at least one reasoning step. This rigorous filtering explicitly rejects non-visually grounded hallucinations, ensuring the distilled trajectories actively rely on relevant visual evidence. The detailed procedure is outlined in Algorithm~\ref{alg:bbox_filtering}.

\begin{algorithm}[h]
\caption{Attention-BBox Overlap Filtering}
\label{alg:bbox_filtering}
\begin{algorithmic}[1]
\REQUIRE Dataset $\mathcal{D}$, VLM model $\mathcal{M}$, Acceptance threshold $\tau = 0.20$, Steps $K$
\ENSURE Filtered dataset $\mathcal{D}_{accept}$
\STATE $\mathcal{D}_{accept} \leftarrow \emptyset$
\FOR{\textbf{each} sample $s \in \mathcal{D}$}
    \STATE $I \leftarrow s.\text{image}, Q \leftarrow s.\text{question}, B \leftarrow s.\text{gt\_bboxes}$
    \STATE $A_{steps} \leftarrow \text{ExtractAttention}(\mathcal{M}, I, Q, K)$
    \STATE $accepted \leftarrow \text{False}$
    \FOR{$t = 1$ \TO $K$}
        \STATE $a \leftarrow A_{steps}[t]$
        \STATE $mask \leftarrow \text{BuildBBoxMask}(B, N_{img}, \text{patch\_grid\_of}(I))$
        \STATE $inside\_weight \leftarrow \sum_i (a[i] \times mask[i])$
        \STATE $total\_weight \leftarrow \sum_i a[i]$
        \IF{$total\_weight > 0$}
            \STATE $ratio \leftarrow inside\_weight / total\_weight$
        \ELSE
            \STATE $ratio \leftarrow 0$
        \ENDIF
        \IF{$ratio \geq \tau$}
            \STATE $accepted \leftarrow \text{True}$
            \STATE \textbf{break}
        \ENDIF
    \ENDFOR
    \IF{$accepted = \text{True}$}
        \STATE $\mathcal{D}_{accept} \leftarrow \mathcal{D}_{accept} \cup \{s\}$
    \ENDIF
\ENDFOR
\RETURN $\mathcal{D}_{accept}$
\end{algorithmic}
\end{algorithm}

We constructed \textbf{LaViT-SFT-15k}, a high-quality multimodal dataset specifically designed to train the latent visual thought capabilities of the model. The dataset comprises \textbf{14,567} samples, aggregating diverse visual scenarios from 10 mainstream vision-language benchmarks.

\paragraph{Data Distribution.}
The dataset ensures diversity by incorporating samples from captioning, VQA, and document understanding tasks. As shown in Table~\ref{tab:data_distribution}, \texttt{Flickr30k} (32.13\%) and \texttt{GQA} (20.05\%) constitute the majority of the data, providing a strong foundation for general visual grounding, while datasets like \texttt{DocVQA} and \texttt{TextCap} enhance the model's fine-grained perception capabilities.

\begin{table}[h]
\centering
\small
\caption{Distribution of data sources in LaViT-SFT-15k.}
\label{tab:data_distribution}
\begin{tabular}{lrr}
\toprule
\textbf{Source Dataset} & \textbf{Samples} & \textbf{Percentage} \\
\midrule
Flickr30k & 4,681 & 32.13\% \\
GQA & 2,921 & 20.05\% \\
DocVQA & 1,647 & 11.31\% \\
TextCap & 1,383 & 9.50\% \\
Visual7W (V7W) & 1,110 & 7.62\% \\
OpenImages & 1,016 & 6.97\% \\
TextVQA & 839 & 5.76\% \\
InfographicsVQA & 651 & 4.47\% \\
CUB (Birds) & 174 & 1.19\% \\
VSR & 145 & 1.00\% \\
\midrule
\textbf{Total} & \textbf{14,567} & \textbf{100.00\%} \\
\bottomrule
\end{tabular}
\end{table}

\paragraph{Statistical Properties.}
Table~\ref{tab:stat_properties} summarizes the key statistical properties of the text and visual components. The dataset features high-resolution inputs (avg. $957 \times 882$) and rich visual semantics, represented by an average of 697.73 visual tokens per image. The high-dimensional visual features ($D=5120$) serve as the supervision target for the latent thought process.

\begin{table}[h]
\centering
\small
\caption{Statistical properties of textual and visual components in LaViT-SFT-15k.}
\label{tab:stat_properties}
\begin{tabular}{lc}
\toprule
\textbf{Statistic} & \textbf{Value} \\
\midrule
\multicolumn{2}{l}{\textit{Textual Statistics}} \\
Avg. Question Length & 13.58 words \\
Max Question Length & 34 words \\
Avg. Answer Length & 4.92 words \\
\midrule
\multicolumn{2}{l}{\textit{Visual Statistics}} \\
Avg. Image Resolution & $957 \times 882$ pixels \\
Avg. Visual Tokens & 697.73 \\
Feature Dimension & 5,120 \\
\bottomrule
\end{tabular}
\end{table}

\paragraph{Data Sampling and Filtration.}
To construct the high-quality LaViT-SFT-15k dataset, we utilized the VisCot dataset as our foundational sampling pool. As detailed in Table~\ref{tab:data_filtration}, we traversed a total of 30,500 raw candidate samples during the data collection and alignment pipeline. To ensure precise synchronization between the visual inputs and teacher supervision signals, we applied a rigorous filtration mechanism, resulting in 15,933 samples being discarded and yielding a final curated subset of 14,567 accepted instances (an overall acceptance rate of 47.8\%). 

Further analysis of Table~\ref{tab:data_filtration} reveals that the acceptance rate varies significantly across different source datasets, heavily depending on the difficulty of semantic alignment. Datasets requiring fine-grained visual grounding or dense object perception exhibited substantial rejection rates. For instance, CUB and OpenImages recorded the lowest acceptance rates of 30.4\% and 36.0\%, respectively, while Flickr30k saw over 6,000 candidates filtered out. In contrast, structure-rich and text-intensive datasets demonstrated much higher retention rates; TextVQA and TextCap achieved impressive acceptance rates of 74.6\% and 66.1\%, respectively. This selective sampling strategy guarantees that our final dataset is densely packed with high-fidelity, complex reasoning scenarios while strictly controlling alignment noise.

\begin{table}[h]
\centering
\small
\caption{Statistics of data sampling and filtration for each source dataset in LaViT-SFT-15k. The base sampling pool is constructed from the VisCot dataset. `Total Candidates' is the sum of accepted and filtered samples.}
\label{tab:data_filtration}
\resizebox{\linewidth}{!}{%
\begin{tabular}{lrrrr}
\toprule
\textbf{Source Dataset} & \textbf{Total Candidates} & \textbf{Accepted} & \textbf{Filtered} & \textbf{Accept. Rate} \\
\midrule
Flickr30k & 10,699 & 4,681 & 6,018 & 43.8\% \\
GQA & 6,727 & 2,921 & 3,806 & 43.4\% \\
DocVQA & 3,165 & 1,647 & 1,518 & 52.0\% \\
TextCap & 2,094 & 1,383 & 711 & 66.1\% \\
Visual7W (V7W) & 1,951 & 1,110 & 841 & 56.9\% \\
OpenImages & 2,819 & 1,016 & 1,803 & 36.0\% \\
TextVQA & 1,125 & 839 & 286 & 74.6\% \\
InfographicsVQA & 1,130 & 651 & 479 & 57.6\% \\
CUB (Birds) & 573 & 174 & 399 & 30.4\% \\
VSR & 217 & 145 & 72 & 66.8\% \\
\midrule
\textbf{Total} & \textbf{30,500} & \textbf{14,567} & \textbf{15,933} & \textbf{47.8\%} \\
\bottomrule
\end{tabular}
}
\end{table}

%
%
\bibliographystyle{splncs04}
\bibliography{main}
\end{document}